# Unsupervised Industrial Anomaly Detection via Pattern Generative and Contrastive Networks

Jianfeng Huang, Chenyang Li, Yimin Lin*, and Shiguo Lian*
AI Innovation and Application Center, China Unicom, Beijing, China.
*Corresponding author(s).

***Abstract*—It is hard to collect enough flaw images for training deep learning network in industrial production. Therefore, existing industrial anomaly detection methods prefer to use CNN-based unsupervised detection and localization network to achieve this task. However, these methods always fail when there are varieties happened in new signals since traditional end-to-end networks suffer barriers of fitting nonlinear model in high-dimensional space. Moreover, they have a memory library by clustering the feature of normal images essentially, which cause it is not robust to texture change. To this end, we propose the Vision Transformer based (VIT-based) unsupervised anomaly detection network. It utilizes a hierarchical task learning and human experience to enhance its interpretability. Our network consists of pattern generation and comparison networks. Pattern generation network uses two VIT-based encoder modules to extract the feature of two consecutive image patches, then uses VIT-based decoder module to learn the human designed style of these features and predict the third image patch. After this, we use the Siamese-based network to compute the similarity of the generation image patch and original image patch. Finally, we refine the anomaly localization by the bi-directional inference strategy. Comparison experiments on public dataset MVTec dataset show our method achieves 99.8% AUC, which surpasses previous state-of-the-art methods. In addition, we give a qualitative illustration on our own leather and cloth datasets. The accurate segment results strongly prove the accuracy of our method in anomaly detection.***

## I. INTRODUCTION

IN industrial production, there may produce anomaly on the product for production technology of high temperature or high pressure. The anomaly always appears in a low frequency but has a serious impact on product quality. In recent years, a growing number of neural networks were designed for defect detection and achieved extraordinary performance. Existing deep learning methods for anomaly detection can be divided into supervised learning [1, 2, 3, 4] and unsupervised learning [5, 6, 7].

Jianfeng Huang, Chenyang Li, Yimin Lin and Shiguo Lian are with China United Communications Co. , Ltd., 51 Hepingli West St, Dongcheng District, Beijing, China (e-mail: {huangjf53,licy1039,linym39,liansg}@chinaunicom.cn).

Supervised learning network needs to give the category and location of anomaly, which can be solved with a number and bounding box [8]. Different number corresponds to different category, while the bounding box locates the anomaly. To achieve flaw detection, the first thing is to collect enough anomaly images and label the category and location of anomaly with bounding box. After that we can train many kinds of object detection networks with these label images. For example, Faster RCNN[1] or SPP-Net [2] firstly selects the candidate area of anomaly and then exactly locates the location of anomaly in the candidate area. Different from these methods, some networks like yolo [3] or SSD [4] doesn't need regional proposal and gives the final detection result directly. However, for all the supervised learning networks, enough anomaly samples are necessary for a great performance. Unfortunately, in most practical production scenarios, the anomaly appears in a low frequency. Therefore, it is hard to apply supervised learning network in actual application.

To solve the absence of anomaly sample, some research institutes proposed unsupervised anomaly detection methods. These unsupervised anomaly detection networks can build a detection model without any annotated samples, which makes it is very suitable for scenarios described above. Following this thought, large numbers of unsupervised methods were proposed. Some researchers solve this problem with reconstruction thought [5, 6]. The core idea is to conduct encoding and decoding on the input normal images and train the neural network with the aim of reconstruction. However, the networks above build a memory library by clustering normal images, which lead to a poor expansibility when there comes a new kind of texture.

In fact, traditional end-to-end networks suffer barriers of fitting nonlinear model in high-dimensional space especially when the training data is limited. Inspired by human always learn to remember the image they have seen before and find the difference between anomaly image and their imagine. Therefore, we also use a hierarchical task learning that include pattern generation and comparison stages. The pattern generation network tries to learn the regulation of texture and generate an accurate and reliable image, which makes it robust to texture change. Here we select the state-of-the-art VIT-based network since it pays more attention to global feature, which makes the generative image have an excellent performance on recovering the texture details. The next pattern comparison network tries to find out the differences and locate the anomaly position by a Siamese network [9, 10]. We observe that the prediction becomes wider especially on the edge of defect due to the inferring direction. Finally, we present a bi-directional inference and take intersection operation of their predictions in order to improve the accuracy of anomaly localization.

To summarize, our paper has following contributions:

1. We propose a combination of generation and comparison networks according to human experience. Our network can solve the problem that is existing end-to-end network needs extensive training image data in industrial quality inspection.

2. We propose a VIT-based encoder-decoder pattern generation network which is robust to texture change. We firstly classify the image data into several categories according to typical human design style, such as the size or periodicity of texture, and then use these data to train our network, which makes our network can infer correct results even though there is texture change.

3. We propose a comparison network by increasing up-sample layer in the decision module of traditional Siamese network. After this, our comparison network can not only output the similarity of two input image, but also show the precise difference location with a mask image.

4. We use the intersection of two direction inference results to improve anomaly detection precision, and achieve best performance on public MVTec AD dataset comparing to the state-of-the-art methods.

The rest of the paper is organized as follows: in Section II we present an overview for unsupervised learning approaches that have been presented for industrial anomaly detection and describe some related networks. In Section III, we present the specific network used in this work. Experiments presented on MVTec and our own dataset in Section IV prove the excellent performance of our network for the anomaly detection task. Section V is the conclusion.

## II. RELATED WORK

### A. Unsupervised Anomaly Detection Methods

As mentioned above, it is hard to collect enough anomaly samples in actual production, so unsupervised anomaly detection methods are suitable for industry anomaly detection. Existing unsupervised anomaly detection methods can be summarized as normalizing flow based and reconstruction based methods. Among normalizing flow based methods, FastFlow [7] expands the original normalizing flow model to two-dimensional space to address the concerns, which achieves an excellent performance. Reconstruction based method UTAD [5] proposes an unsupervised anomaly detection scheme for natural images by combining mutual information, Generative Adversarial Networks [11], and autoencoder. A two-stage framework is utilized to generate high-fidelity and anomaly-free input reconstructions for anomaly detection. Our image generation network is essentially an improved reconstruction method. However, unlike reconstruction method above using CNN as basic network unit in encoder and decoder, we introduce Swin Transformer block [12] instead.

### B.VIT-based Network

Self-attention is proposed by [13], which was used in Natural Language Processing (NLP) at first. For each word of a sentence, self-attention utilizes attention mechanism to compute the association score between this word and the rest. With the association score, self-attention can exactly judge the mean of each word in the sentence. Therefore, it has become the state-of-the-art model in machine translation task. Inspired by self-attention in NLP, [14] introduces self-attention layer in the ResNet and it replaces some of the convention layer with self-attention layers, and conducts self-attention computation within each pixel of a local window. The fusion of self-attention layers makes these methods get a better accuracy on public dataset. However, a large number of memory access leads to poor efficiency in speed. To solve this problem, Liu Z, et al [12] proposes a shifted window based self-attention, which executes self-attention computation in a smaller window. Besides, [12] shifts the windows to achieve information communication in different transform layer. By these operators, [12] has an excellent performance in accuracy and speed. Therefore, we directly use Swin Transformer block as the basic network unit of encoder and decoder module for a better result.

### C. Siamese Network

Image similarity comparison is a very important and fundamental issue in computer vision, which is widely used in image retrieval, object tracking and anomaly detection. Traditional image similarity comparison methods use SIFT [15] feature point to achieve its robustness to illumination, rotation and shift. However, the accuracy of sift is not very satisfying. As the development of deep learning, Siamese network [10] was proposed to solve image similarity comparison issue. Siamese network is made up of feature extraction and decision modules, and feature extraction module usually has two branches to respectively exact feature from two input images. Siamese network can be further divided into Siamese and Pesudo-siamese network according to whether two branches of feature extraction module share weights. The feature extraction module of Siamese network shares weight while Pesudo-siamese network doesn't. Therefore, the weight of Pesudo-siamese network is almost twice than Siamese network, which make it more flexible to extract feature. In our work, we use improved Pesudo-siamese network as our image patch comparison module, which can output similarity and mask simultaneously.

## III. THE PROPOSED METHOD

The architecture of our network is shown in Fig. 1. Our network is mainly composed of image generation and similarity comparison network. Image generation network is basically an VIT-based encoder-decoder module, which can predict the third image patches with the first two adjacent image patches as input. The Similarity Comparison Network is a Siamese Network [10], which can compute the similarity of two image patches and output the corresponding mask image. To avoid flaw edge expansion, we infer from two directions, which are from left to right and from right to left. For each direction, we can get a probability map and mask image. By comparing the probability map with threshold, we get the value in classification map when probability value is less than threshold is set to 1, otherwise, the value is 0. Value 1 and 0 are respectively abnormal and normal label. Finally, we use the intersection of two direction classification maps as final classification map. Similarly, the final mask image is the intersection of two direction mask images.

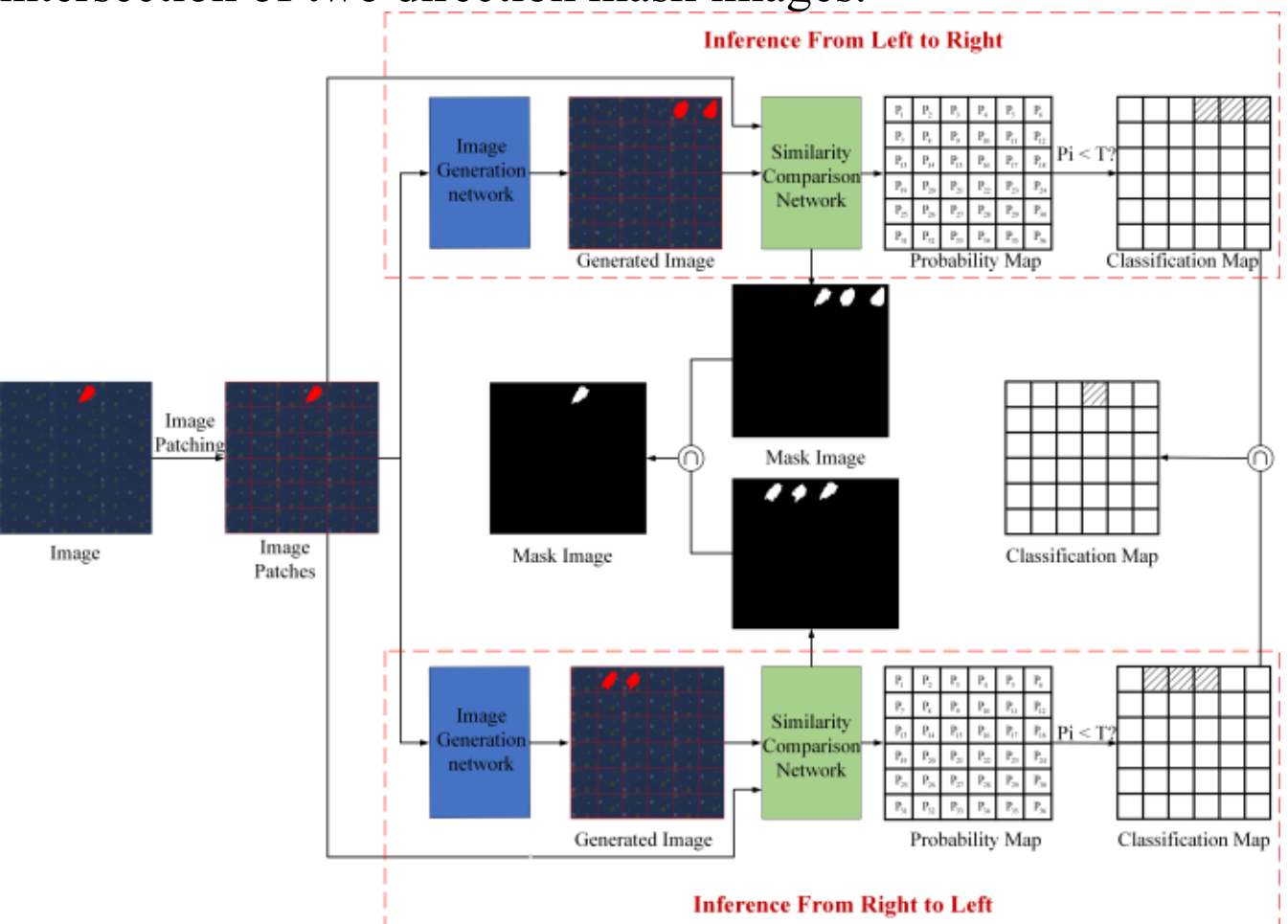

Fig. 1. The overall architecture of our network. It mainly includes two parts, Image Generation Network and Similarity Comparison Network. Image Generation Network is used to predict image patch with two image patches as input, Similarity Comparison Network is used to compute the similarity between two input images.

We will describe the implementation details of these two networks in Section 3.1 and 3.2 respectively.

### A. VIT-based Generation Network

As shown in Fig. 2, the pattern generation network consists of four modules, which are patch partition module, encoder module, decoder module and fusion module, and we will give a particular explanation on these four modules.

Firstly, we use patch partition module to split an input RGB image into N×N image patches, and N is changeable for different tasks. In Fig. 2, the value of N is set to 4. Similar to [12], we use a convolution layer to achieve patch partition shown in Fig. 2. After patch partition, we split the image into 4×4 image patches, and the size of each image patch is (H/4)×(W/4)×3. By feeding two adjacent image patches to next encoder, decoder and fusion modules, we can generate the third adjacent image patch. For instance, we mask 16 image patches with mark 1 to 16 by order. Putting image patches 1 and 2 to the generation network, we can get the predicted image patch 3'. Repeat this operation for all adjacent image patches, we can generate 16 image patches.

Inspired by the encoder and decoder framework of transformer in NLP, we design a decoder module to build association relationship between two input image patches for better prediction result. From Fig. 2, we can see that the decoder module consists of four stages. The input of each decoder stage is the addition or concatenation of corresponding encoder stage and front decoder stage, which can be represented as following formula:

$$\mathrm{DecoderIn}_1 = \mathrm{Encoder1}_4 + \mathrm{Encoder2}_4 \tag{1}$$

$$\mathrm{DecoderIn}_i = \mathrm{Encoder1}_{5-i} + \mathrm{Encoder2}_{5-i} + \mathrm{DecoderOut}_{i-1} \quad (i = 2,3) \tag{2}$$

$$\mathrm{DecoderIn}_4 = \mathrm{Encoder1}_1 \oplus \mathrm{Encoder2}_1 \oplus \mathrm{DecoderOut}_3 \tag{3}$$

Here, $\mathrm{DecoderIn}_i$ means the input of *i-th* decoder stage, while $\mathrm{Decoderout}_i$ is the output of *i-th* decoder stage. $\mathrm{Encoder1}_i$ represent the output of *i-th* stage of Encoder1, similarly, $\mathrm{Encoder2}_i$ represent the output of *i-th* stage of Encoder2. The value of $i$ is from 1 to 4. + corresponds to the additional operator, which does not change the resolution of input feature. ⊕ is concatenation operator, which can increase the channel of feature by concatenating the input feature.

The front three decoder stages have the same architecture, which is made up of patch expanding and several Swin Transformer blocks. In decoder Stage 4, we use a linear project replacing the patch expanding. The linear projection is a conventional layer, which down-sampling the input feature from (H/4)×(W/4)×3C to (H/4)×(W/4)×C. To generate image patch from low dimension feature, we use a patch expanding layer to increase the resolution of low dimension feature like [16]. A patch expanding layer includes two operators, firstly using a linear layer to increase the feature channel to 2× input feature channel, and then expanding the resolution of the input features to 2× original feature by reducing the channels to quarter.

After four decoder stages, we introduce multilayer perceptron (MLP) [17] layer to fuse global and local features. As shown in Fig. 2, we feed the output feature of four decoder stages into MLP layer, whose sizes are, (H/16)×(W/16)×4C, (H/8)×(W/8)×2C, (H/4)×(W/4)×C, (H/4)×(W/4)×C, from Stage 1 to Stage 4. After this, MLP layer outputs a feature with a size of (H/4)×(W/4)×4C. Next, we put this feature to another MLP to predict image patch with a size of (H/4)×(W/4)×3.

### B. Siamese-based Comparison Network

We design our pattern comparison network inspired by Pesudo-siamese network as Fig.3 shown. Similar to Pesudo-siamese, our image patch comparison network is made up of feature extraction module and decision module. Feature extraction module consists of 2 branches, and each includes 3 convolution blocks and 2 max pooling layers. Convolution block is a combination of convolution layer, batch normalization layer (BN) [18] and Relu function. Specifically, the first two convolution blocks don't share weights while the third has same weights. In decision module, the first layer is a concatenate layer, which is used to concatenate the output features of two feature extraction branches. Then we feed the output of concatenate layer into two fully connection layers and obtain the final probability value, which is used to measure the similarity of two input image patches. In addition, the output feature of concatenate layer is fed into two convolution blocks to get the mask image. The first convolution block is also made up of convolution layer, BN and Relu function, while the second use sigmoid function to replace Relu function. Mask image is a gray image, in which pixel value 1 means the difference between two input images while value 0 corresponds to the same. The height and width of mask image is a quarter of the input image patch. According to mask image, we can clearly find the position of difference between two input images. When training the image patch comparison network, we optimize the parameters by minimizing the following loss:

$$\mathcal{L} = \frac{1}{1+e^{\|I_o - I_g\|_1}} + \max(0, 1-y*P) \tag{4}$$

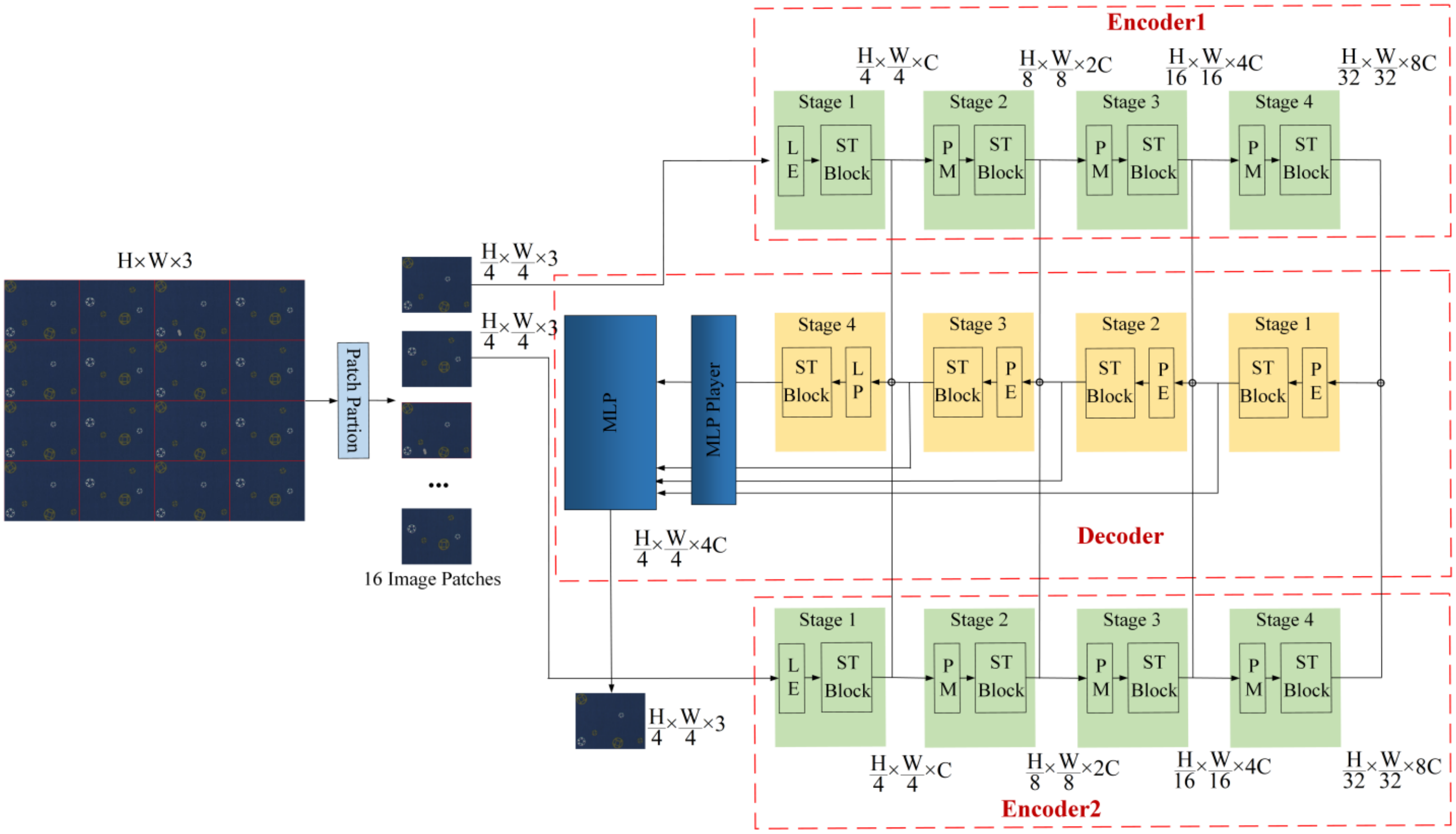

Fig. 2. The architecture of our network. To be more distinct, we use the abbreviation of basic network unit in the architecture figure, where LE means linear embedding, ST block means Swin Transformer block, PM means patch merging, PE means patch expanding, and LP means linear project.

Where $I_o$ is the output mask image, while $I_g$ is the corresponding ground-truth mask image. P is the probability value obtained from the image patch comparison network, and $y \in \{-1, 1\}$ is the corresponding label (with $-1$ and 1 denoting a non-matching and a matching pair, respectively).

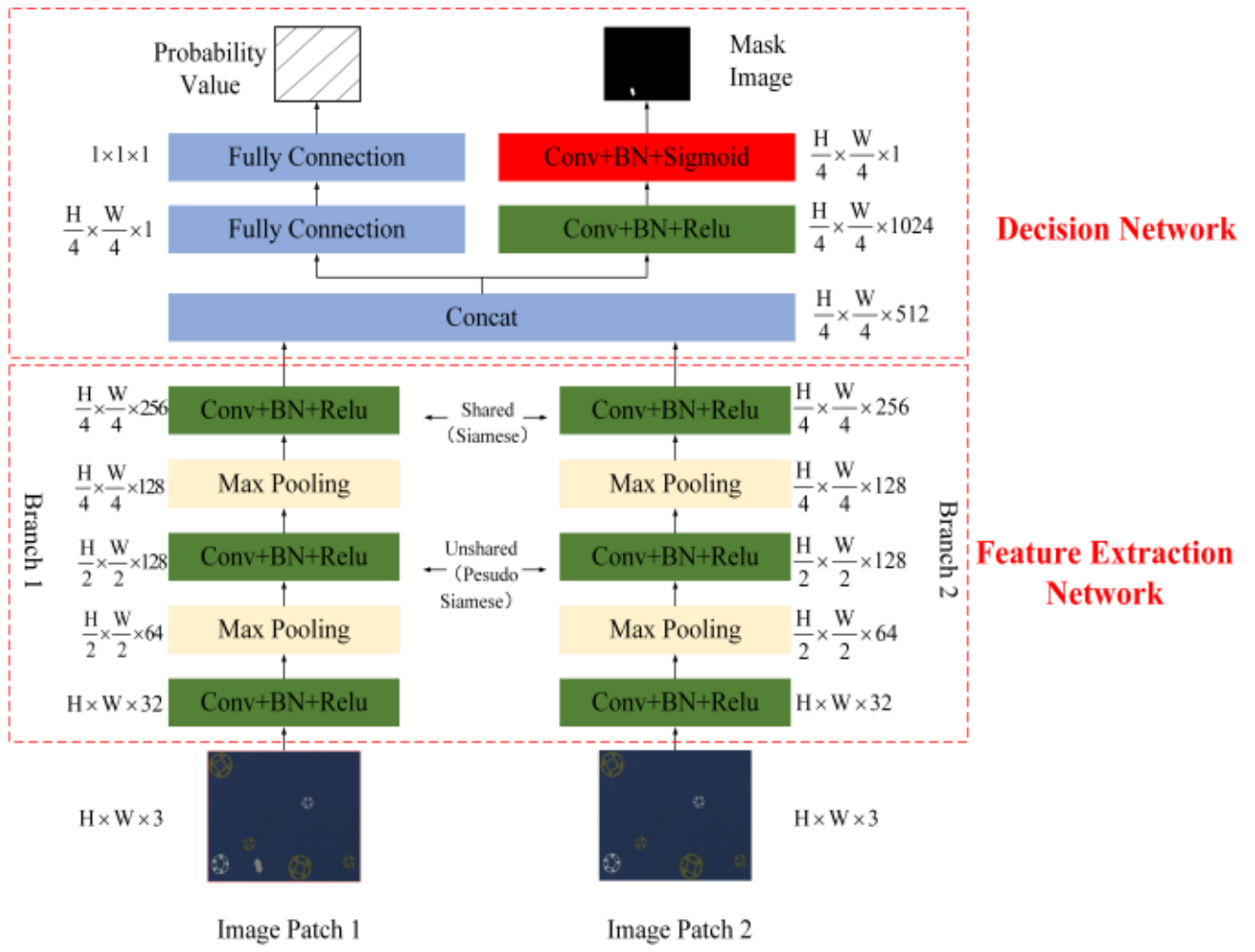

Fig. 3. Framework of image patch comparison network

### C. Bi-directional Inference Strategy

To avoid the anomaly edge was expended, we generate image patch from two directions, which respectively are inference from left to right and inference from right to left. Fig. 4 gives a detail explanation. In Fig. 4, there exists anomaly in $I_4$ image patch of input image. By inferring from left to right, we get left classification map, in which $LC_4$, $LC_5$ and $LC_6$ are abnormal, while the truth is that only $LC_4$ should be abnormal. The reason is that we rebuild $I_4$ with normal image patch $I_2$ and $I_3$ as input, so the generated image patch $I_4$' is normal. Comparing $I_4$' with anomaly image patch $I_4$, the classification result of $LC_4$ is abnormal. However, when rebuild $I_5$' and $I_6$', we respectively use ($I_3$, $I_4$) and ($I_4$, $I_5$) as input. These two groups of input both include anomaly image patch $I_4$, so the generated image patch $I_5$' and $I_6$' are anomaly image patches, which are different from original normal image patch $I_5$ and $I_6$. Therefore, the classification result of $LC_5$ and $LC_6$ are both abnormal. To get precise result, we add new inference from right to left and get right classification map. In right classification map, we can see that the classification result of $RC_2$, $RC_3$ and $RC_4$ are abnormal, but only $RC_4$ is abnormal in fact. The reason for this problem is that we respectively rebuild $I_2$' and $I_3$' with ($I_3$, $I_4$) and ($I_4$, $I_5$) as input, which both include anomaly image patch $I_4$. Therefore, the generated image patch $I_2$' and $I_3$' are anomaly image patch, while the original $I_2$ and $I_3$ are normal. As a summary, there always exists anomaly expansion along the inference direction, so we use the intersection of two inference direction classification maps as our final result, which can be formulated as follow:

$$C_i = LC_i \cap RC_i \tag{5}$$

Where $C_i$ is the final classification result of *ith* image patch, $LC_i$ is left classification result of *ith* image patch by inferring from left to right, $RC_i$ is right classification map of *ith* image patch by inferring from right to left, the range of *i* is from 1 to 36 in this example, $\cap$ means intersection symbol.

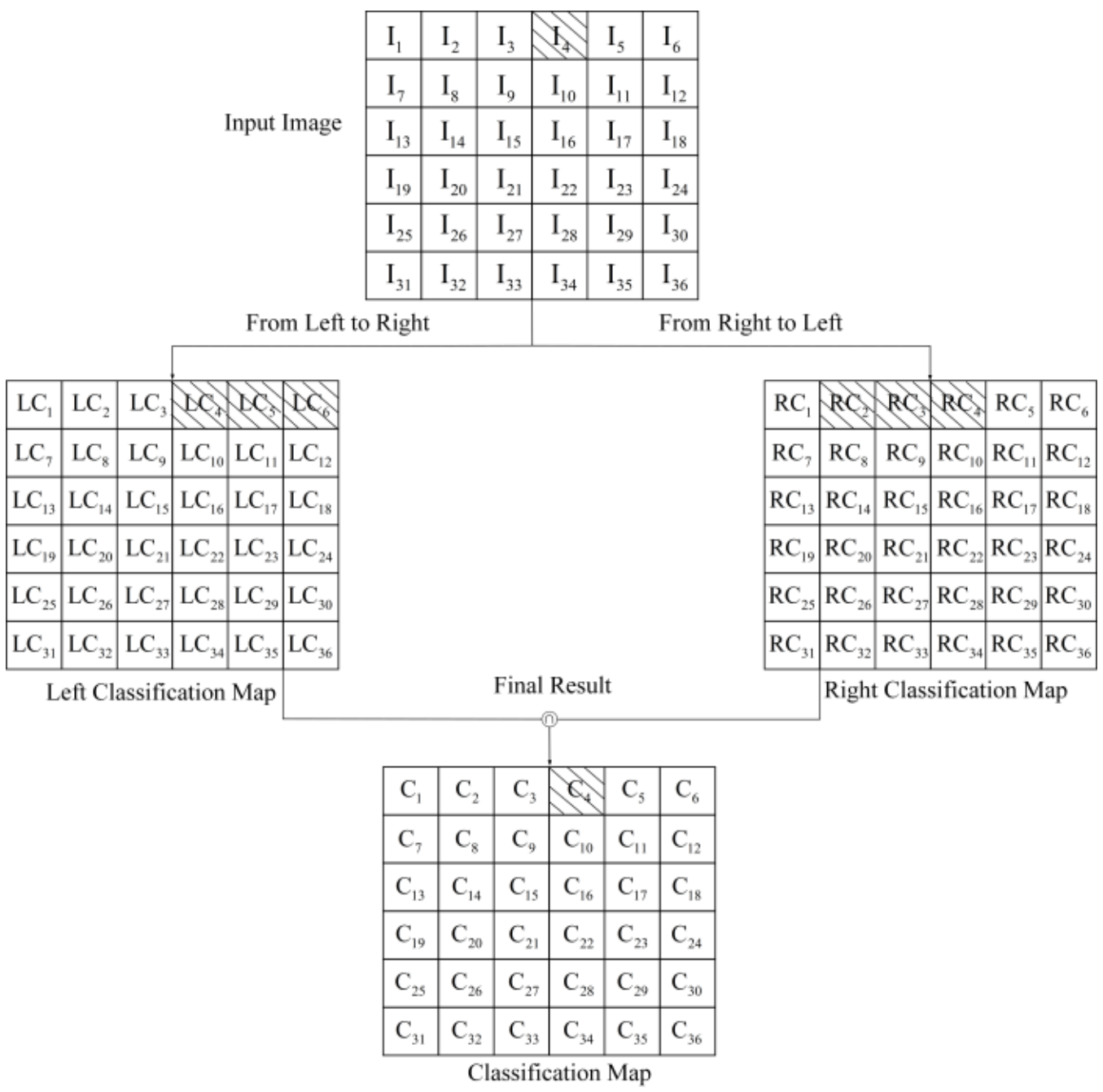

Fig.4. Inference from two directions

## IV. EXPERIMENT

### A. Datasets and Metrics

We evaluate our method on public dataset MVTec AD [19] and our own dataset. MVTec AD is proposed by MVtec Company for industrial anomaly detection. It contains 5354 images. There are 1888 anomaly images with 73 kinds of defects. For each anomaly image, they provide pixel level label. Our own dataset is made up of two kinds of material in total. One is leather and the other is cloth. The leather dataset includes 6 kinds of textures and has 4020 images, among them 2000 flaw images. There are four kinds of flaws and we label all the flaw images at pixel level. The image resolution is 640 * 512. For cloth dataset, there are 5000 images which can be divided into 10 kinds of decorative pattern materials according to typical human designed style. The flaw images account for 30%. Similar to leather dataset, we provide pixel-level label for each flaw cloth image. The main difference between leather and cloth dataset is that the texture of leather is small and repetitive while cloth is big and complex.

We evaluate the accuracy of our method from two scales, image level and pixel level. Similar to FastFlow [7], we use the area under the receiver operating characteristic curve (AUROC) to measure our method.

### B. Implementation Details

**Network details:** For Swin Transformer blocks in encoder and decoder module, we build our base model inspired by Swin Transformer [12], whose windows is set to M = 7. The channel number of the hidden layers in the first stage is set to C = 24, and layers number of encoder and decoder are 2, 2, 6, and 2 in four stages.

**Train details:** We train our network in two steps. Firstly, we train our pattern generation network. Next, we fix the weight of pattern generation network and train the image patch comparison network. Inspired by [20], we train our image patch generation network with random mask method on ImageNet [21] and our own dataset. For each training image, firstly we randomly select three adjacent image patches. Then we respectively use left or right two adjacent image patches as input, and generate the corresponding third image patch. Similar to [20], we use the mean squared error as loss function as follow:

$$\mathrm{MSE}(\mathrm{I}_1, \mathrm{I}_2) = \frac{1}{\mathrm{MN}} \sum_{i=1}^{N} \sum_{j=1}^{M} \| \mathrm{I}_1(i, j) - \mathrm{I}_2(i, j) \|^2 \tag{6}$$

Where $I_1$ is the pattern generated by image patch generation network, while $I_2$ is the corresponding ground truth. M, N are the width and height of image patch. By adjusting weights to make the MSE value as small as possible, we can get the final image patch generation model.

### C. Evaluation on MVTec AD Dataset

**Quantitative Results:** We give a quantitative comparison between our proposed method and other state-of-the-art anomaly detection works on MVTec AD dataset. MVTec AD dataset consists of 15 industrial products, among which 10 are objects and 5 are textures. For each industrial product, there are only normal images as training set while the mixture of normal and flaw images as test set. We use image level AUC and pix level AUC as the metric standard to compare our proposed method with other methods, including PatchSVDD [22], SPADE* [23], DifferNet [24], PaDim [25], Cut Past[26], Patchcore [27], CFlow[28], and Fastflow[7]. The comparison result is shown in Table1. From Table 1, we can see that our method achieves 99.8 on image level AUC and 98.7 on pixel level AUC, which is the best in all anomaly detection tasks.

**Qualitative Results:** We give a qualitative comparison between our proposed method and PatchCore on MVTec AD dataset, and the result is shown in Fig. 5. We can see that the heatmaps of PatchCore cannot get an accurate segment mask when the flaw of input image is not obvious such as screw, toothbrush and carpet. For the product with obvious flaw like leather, title and wood, PatchCore can only judge that there is flaw in image, but cannot segment the shape of flaw. In contrast, our method can give accurate heat mask for all the flaws. Besides, as we can see in second group, there are false-alarms existing in PatchCore. Meanwhile, our method can exactly judge that these are a normal image.

TABLE I
COMPARED WITH OTHER METHODS ON MVTEC AD DATASET

| Method | PatchSVDD | SPADE* | DifferNet | PaDim | Cut Past | Patchcore | CFlow | Fastflow | Ours |
|---|---|---|---|---|---|---|---|---|---|
| carpet | (92.9,92.6) | (98.6,97.5) | (84.0, -) | (-,99.1) | (**100.0**,98.3) | (98.7,98.9) | (**100.0**,99.3) | (**100.0**,99.4) | (**100.0**,**99.5**) |
| grid | (94.6,96.2) | (99.0,93.7) | (97.1, -) | (-, 97.3) | (96.2,97.5) | (98.2,98.7) | (97.6,**99.0**) | (99.7,98.3) | (**100.0**,98.5) |
| leather | (90.9,97.4) | (99.5,97.6) | (99.4, -) | (-, 99.2) | (95.4,99.5) | (**100.0**,99.3) | (97.7,**99.7**) | (**100.0**,99.5) | (99.8,98.6) |
| tile | (97.8,91.4) | (89.8,87.4) | (92.9, -) | (-, 94.1) | (**100.0**,90.5) | (98.7,95.6) | (98.7,98.0) | (**100.0**,96.3) | (**100.0**,**98.5**) |
| wood | (96.5,90.8) | (95.8,88.5) | (99.8, -) | (-, 94.9) | (99.1,95.5) | (99.2,95.0) | (99.6,96.7) | (**100.0**,97.0) | (**100.0**,**99.1**) |
| bottle | (98.6,98.1) | (98.1,98.4) | (99.0, -) | (-, 98.3) | (99.9, 97.6) | (**100.0**,98.6) | (**100.0**,**99.0**) | (**100.0**,97.7) | (**100.0**,98.4) |
| cable | (90.3,96.8) | (93.2,97.2) | (86.9, -) | (-, 96.7) | (**100.0**,90.0) | (99.5,98.4) | (**100.0**,97.6) | (**100.0**,98.4) | (99.1,**98.6**) |
| capsule | (76.7,95.8) | (98.6,99.0) | (88.8, -) | (-, 98.5) | (98.6,97.4) | (98.1,98.8) | (99.3,99.0) | (**100.0**,**99.1**) | (**100.0**,98.2) |
| hazelnut | (92.0,97.5) | (98.9,99.1) | (99.1, -) | (-, 98.2) | (93.3,97.3) | (**100.0**,98.7) | (96.8,98.9) | (**100.0**,99.1) | (99.7,**99.3**) |
| mata nut | (94.0,98.0) | (96.9,98.1) | (95.1, -) | (-, 97.2) | (86.6,93.1) | (**100.0**,98.4) | (91.9,98.6) | (**100.0**,98.5) | (99.6,**99.5**) |
| pill | (86.1,95.1) | (96.5,96.5) | (95.9, -) | (-, 95.7) | (99.8,95.7) | (96.6,97.1) | (99.9,99.0) | (99.4,**99.2**) | (**100.0**,98.5) |
| screw | (81.3,95.7) | (99.5,98.9) | (99.3, -) | (-, 98.5) | (90.7,96.7) | (98.1,**99.4**) | (**99.7**,98.9) | (97.8,**99.4**) | (99.5,98.6) |
| toothbrush | (100.0,98.1) | (98.9,97.9) | (96.1, -) | (-, 98.8) | (97.5,98.1) | (**100.0**,98.7) | (95.2,**99.0**) | (94.4,98.9) | (**100.0**,98.5) |
| transistor | (91.5,97.0) | (81.0,94.1) | (96.3, -) | (-, 97.5) | (99.8,93.0) | (**100.0**,96.3) | (99.1,98.0) | (99.8,97.3) | (**100.0**,**98.6**) |
| zipper | (97.9,95.1) | (98.8,96.5) | (98.6, -) | (-, 98.5) | (**99.9**,**99.3**) | (98.8,98.8) | (98.5,99.1) | (99.5,98.7) | (99.6,98.6) |
| **AUCROC** | (92.1,95.7) | (96.2,96.5) | (94.9, -) | (97.9,97.5) | (97.1,96.0) | (99.1,98.1) | (98.3,98.6) | (99.4,98.5) | (**99.8**,**98.7**) |

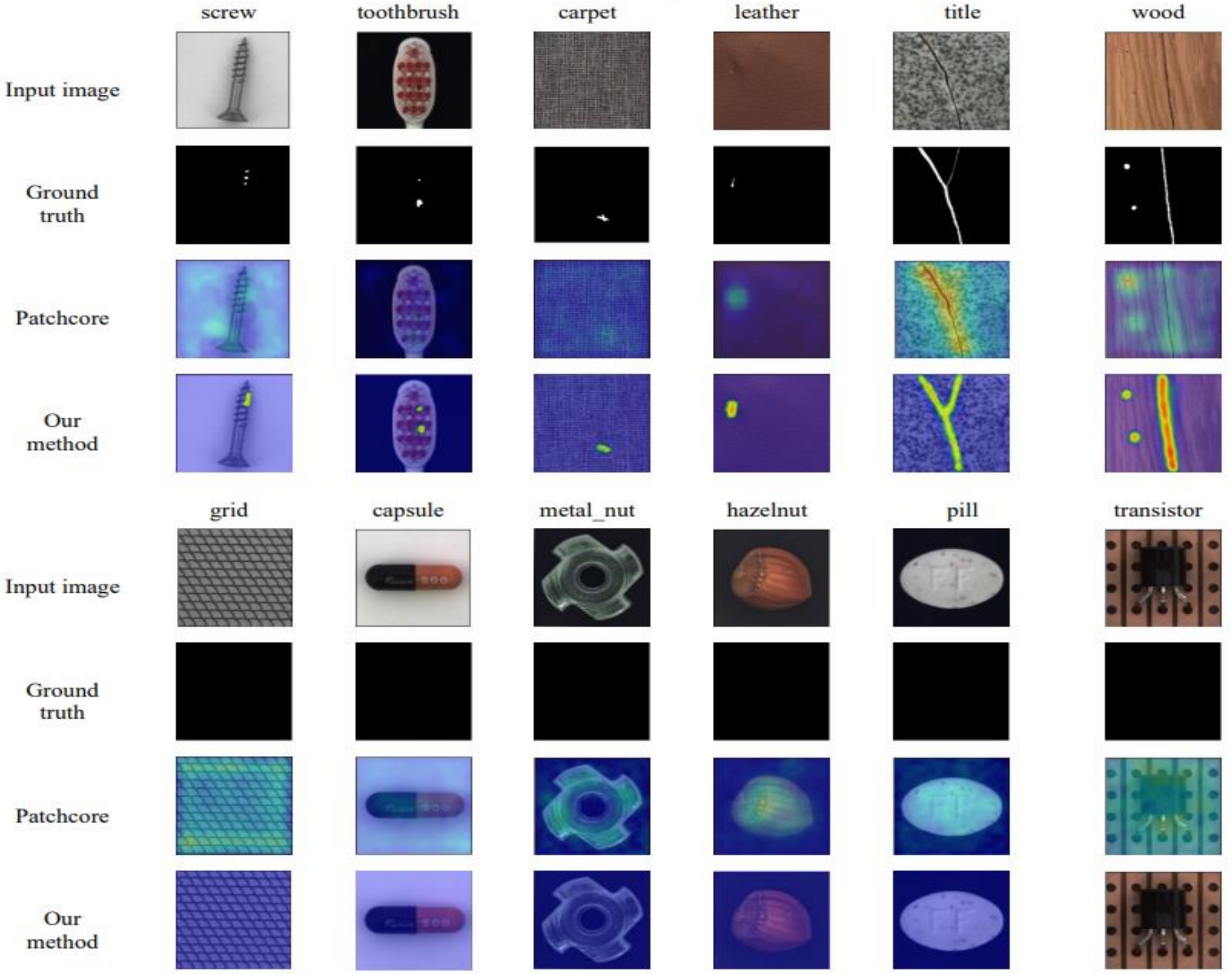

Fig.5. Qualitative Results compared with patchcore on MVTec AD dataset. There are two groups of images. The first group consists of row 1 to 4 with flaw image as input. And the second group is row 5 to 8, in which the input image is normal. For each group, the top row is the input image, and the following row is ground truth mask. The third row is the heatmap predicted by PatchCore. The last row is the mask are placed predicted by our network. As for the column, it represents different industrial products.

## D. Evaluation on Our Dataset

**Qualitative Results:** We also test our network on our own dataset and the qualitative results are shown in Fig. 6. The left three columns are results on leather dataset while the right three are on cloth dataset. Leather dataset has a small and regular texture. Therefore, it is easy to get a wonderful result. From the left of Fig.6, we can see that our method can exactly segment the shape of flaw in leather image even if the flaw is unobvious.

For example the second row is a small hole and the third row is a slight emboss. Cloth dataset includes three kinds of textures, and the complex decorative pattern makes it hard to segment the shape of flaw compared with leather image. Fortunately, from the detection results shown in the right of Fig. 6, our method can get an excellent result even the image texture is more complex. The reason is that our network can learn the repeatable regulation in image, by which it can exactly generate corresponding normal image. Therefore, we can get a wonderful segmentation result after comparing generation image and input image.

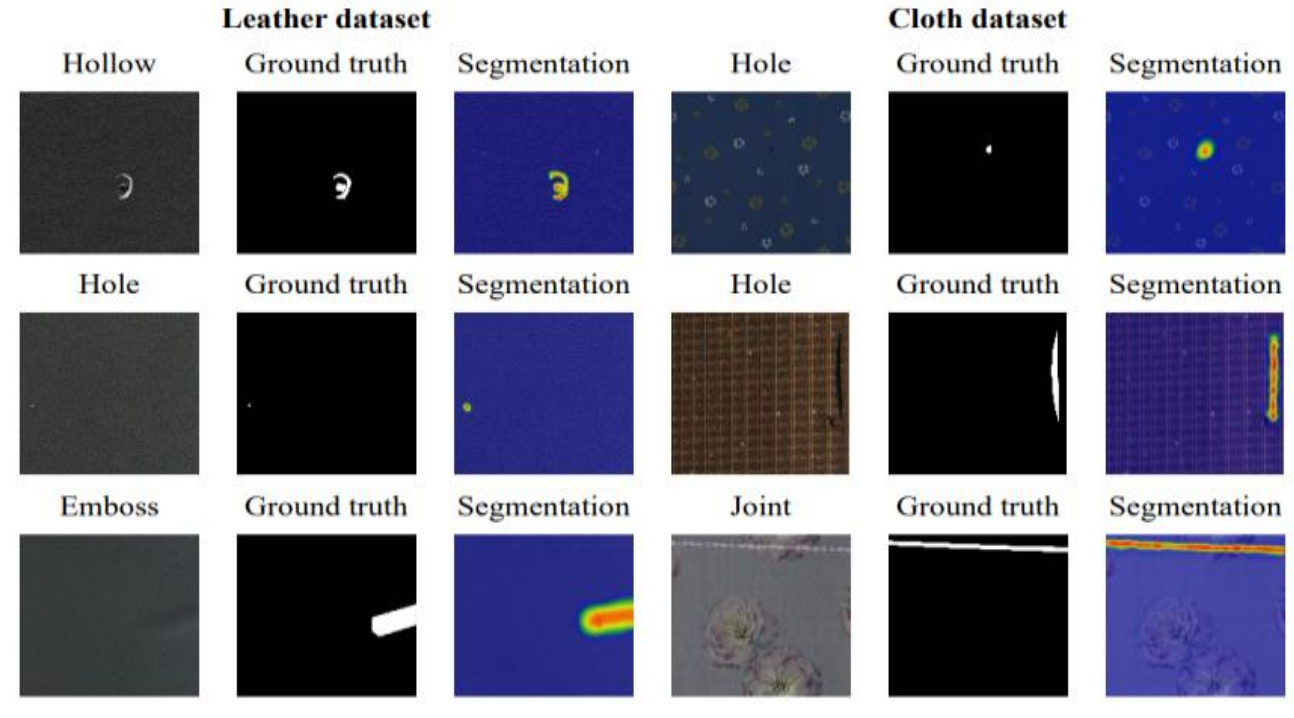

Fig.6. Qualitative Results on Leather and Cloth dataset.

## E. Ablation Study

**Texture robustness test:** we design additional experiments to test the robustness of our method when texture changes. We select two kinds of decorative pattern clothes as Fig. 7 shown. Experiment 1 uses Cloth 1 as the train set, Cloth 2 as the test set. Experiment 2 uses Cloth 2 as the train set, Cloth 2 as the test set. And Table 2 shows the performance of our method and PatchCore on these two groups of experiments.

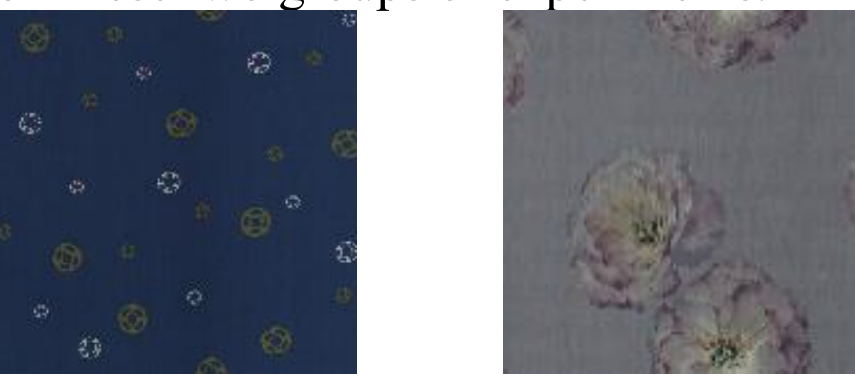

Fig.7 Two kinds of decorative pattern clothes

TABLE 2

THE COMPARISON RESULTS WITH PATHCORE ON TWO KINDS OF DECORATIVE PATTERN CLOTHES

| | Experiment1 | Experiment2 |
|---|---|---|
| PatchCore | (42.2, 67.2) | (92.0, 82.6) |
| Ours | (98.0, 92.0) | (98.0, 92.0) |

From Table 2, we can see that the image level and pixel level AUROC of PatchCore is up to (92, 82.6) when the train set and test set have the same texture. While the AUROC value of PatchCore rapidly decreases to (42.2,67.2) when the train set and test set have different textures. By constrast, our method can keep a stable value of (98, 92) whether the train set and test set are the same or not. The reason is that our method can generate image patch use the learned regularity of texture instead of building a memory bank based on the cluster of normal image sample.

**Influence of image patch**: we investigate the influence of image patch number on cloth image with different decorative pattern. Fig. 8 shows the results. From the AUROC-patch number curve, we can see that Texture 1 achieves the best performance in detection and segmentation when we divide the input image into 4×4, while Texture 2 gets the best detection and segmentation AUROC with number of 7×7. Therefore, when a new texture comes, our inference network need to find out the number of image patch when it get a best performance.

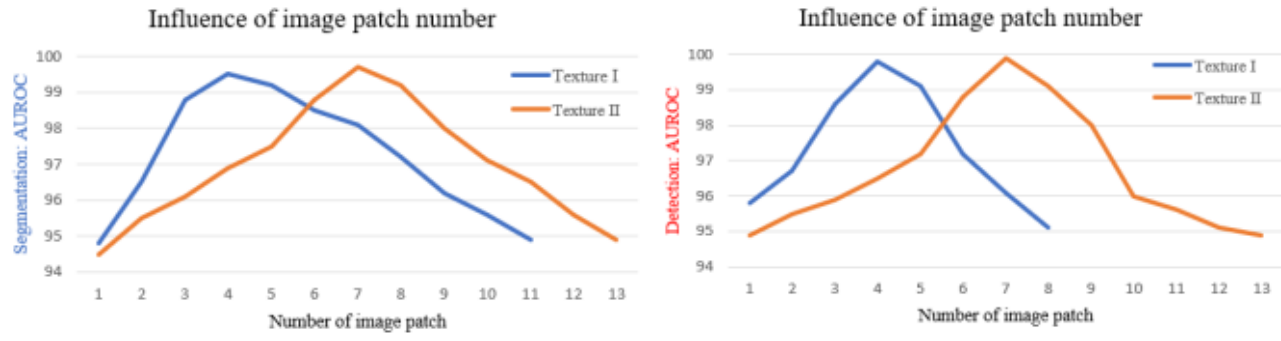

Fig.8 The influence of image patch number, x-axes correspond to the number of image patch, y-axes are respectively segmentation and detection AUROC.

**Performance between CNN and Swin Transformer:** To prove that VIT-based method has the best performance, we also attempt to compare the generation results based on our Swin Transformer encode-decode network and CNN-based Cycle GAN network. And Table 3 shows the detection results of different architecture networks.

TABLE 3

THE DETECTION RESULTS OF DIFFERENT ARCHITECTURE NETWORKS

| | CNN Texture 1 | Ours Texture 1 | CNN Texture 2 | Ours Texture 2 |
|---|---|---|---|---|
| AUCROC | (96.0, 92.9) | (98.0, 95.9) | (92.0, 89.7) | (98.4, 93.9) |

We test these two networks on two kinds of texture cloth dataset. Texture 1 has small decorative pattern as background while Texture 2 has big decorative pattern. We can see that CNN-based detection network is affected by the size of decorative pattern whose results decrease from (96.0, 92.9) to (92.0, 89.7) while our method is robust, (98.0, 95.9) is almost the same as (98.4, 93.9). The reason is that CNN-based detection network has a local receptive field, meanwhile, our VIT-based network can effectively keep the global information of input image, which makes it robust to size changes of any pattern.

## V. CONCLUSION

In this paper, we propose a unsupervised industrial anomaly detection via pattern generative and contrastive networks. Existing CNN-based methods are highly depend on training image samples, which make them work as data driven learning and have poor performance in extensibility. Besides, the characteristic of CNN networks lead to ignore the global feature, which is unfavorable to rebuild the texture detail. Constrastly, our VIT-based encoder and decoder generation framework has an advantage in learning the regulation of texture and rebuilding texture details, which effectively promotes the extensibility on industrial anomaly detection task. We also propose a novel Pesudo-siamese comparsion network to pridect the similarity of the two images and output the probability value and mask image at last. Finally we present a bi-directional inference strategy to improve the performance when the flaw suffers from edge expansion in a single inference. The excellent experiment results on MVTec and our own datasets show our method can obtain better results compared to other state-of-the-art methods on industrial anomaly detection.